# LSTM VS. FEED-FORWARD AUTOENCODERS FOR UNSUPERVISED FAULT DETECTION IN HYDRAULIC PUMPS

**P. Sánchez, K. Reyes, B. Radu, E. Fernández**

Department of Computer Science, University of Alcalá, Spain

## ABSTRACT

Unplanned failures in industrial hydraulic pumps can halt production and incur substantial costs. We explore two unsupervised autoencoder (AE) schemes for early fault detection: a feed-forward model that analyses individual sensor snapshots and a Long Short-Term Memory (LSTM) model that captures short temporal windows. Both networks are trained only on healthy data drawn from a minute-level log of 52 sensor channels; evaluation uses a separate set that contains seven annotated fault intervals. Despite the absence of fault samples during training, the models achieve high reliability (precision ∼ 0.93, recall ≥ 0.99).

**Keywords:** predictive maintenance, anomaly detection, LSTM autoencoder, hydraulic pumps, unsupervised learning

## I. INTRODUCTION

Hydraulic pumps are indispensable in a wide spectrum of industrial processes, from manufacturing presses to hydraulic power units in renewable-energy plants. A single unexpected pump failure can shut down an entire production line, incur costly downtime, and compromise safety. Conventional maintenance relies on either fixed maintenance intervals or supervised Remaining Useful Life (RUL) models that require run-to-failure data. Such supervised approaches typically include regression-based models (e.g., linear or regularized regression), classification techniques (e.g., Logistic Regression, Support Vector Machines, Random Forests), or stochastic degradation models, all of which assume the availability of labelled failure or degradation trajectories.

In practice, however, preventive policies usually replace or repair a pump long before catastrophic failure, so fully labelled degradation trajectories are scarce (Tziolas et al., 2022). This data limitation is well documented in predictive maintenance literature and significantly constrains the applicability of supervised learning techniques in real industrial settings (Carvalho et al., 2019). Unsupervised anomaly detection provides a pragmatic alternative for predictive maintenance in such data-scarce settings (Carrasco et al., 2021). Instead of predicting a precise life estimate, the algorithm learns what healthy operation looks like and raises an alert whenever new sensor readings deviate from that baseline. This paradigm contrasts with supervised classifiers and regressors, which explicitly learn decision boundaries or degradation mappings from labelled fault data, and is particularly appealing when faults are rare, heterogeneous, or costly to label. AEs (which are neural networks trained to reproduce their input) are particularly attractive for this task because they compress complex patterns into a low-dimensional latent space. If the latent code faithfully reconstructs healthy data, anomalous data should yield large reconstruction errors, offering a simple and interpretable anomaly score (Tziolas et al., 2022). This reconstruction-based principle has been widely adopted in industrial anomaly detection, where AEs often serve as unsupervised baselines against more complex deep architectures or hybrid models (Sakurada et al., 2014; Qian et al., 2022).

This work evaluates two AE variants on a real hydraulic-pump dataset: a feed-forward dense AE that operates on single time steps and a LSTM-AE that models five-minute sliding windows. While dense AEs resemble classical snapshot-based models that ignore temporal dependence, recurrent architectures such as LSTMs explicitly capture sequential dynamics, which have been shown to outperform static models when degradation evolves gradually or manifests as transient patterns. The comparison highlights whether explicit temporal modelling confers tangible benefits over a simpler snapshot-based approach when only healthy data are available for training.

## II. ANOMALY DETECTION APPROACH

AE-driven anomaly detection has gained traction across many condition-monitoring domains, including rotating machinery (Ahmad et al., 2020), turbofan engines (Jakubowski et al., 2021), and hydraulic systems (Fic et al., 2023). Most studies follow a similar recipe: train an AE on healthy data, compute reconstruction error at inference time, and flag large errors as faults. The main design choices revolve around (i) the encoder–decoder architecture, (ii) the loss function, and (iii) the thresholding strategy.

Early work employed shallow feed-forward AEs, which capture nonlinear correlations but ignore temporal context. Subsequent research introduced recurrent AEs like LSTM or Gated Recurrent Unit (GRU), which are able to model sequential dependence (Guo et al., 2018; Said Elsayed et al., 2020), consistently outperforming snapshot models when faults develop gradually or manifest as transient drifts. This advantage mirrors findings from supervised time-series models, where LSTM-based architectures outperform classical statistical models such as Autoregressive Integrated Moving Average (ARIMA) under nonstationary operating conditions (Taslim & Murwantara, 2024). Alternative approaches include variational AEs for probabilistic scoring, convolutional AEs that exploit local structure in high-frequency vibration data, and adversarial AEs that learn richer latent spaces at the cost of more complex training (Zenati et al., 2018).

Threshold selection is another active topic (Herrmann et al., 2024). Fixed thresholds based on empirical quantiles are simple but may drift as operating conditions evolve. Adaptive thresholds such as exponentially weighted moving averages, or statistical tests based on the Generalised Extreme Value (GEV) distribution have been proposed to mitigate this issue. Related ideas also appear in classical statistical process control and distance-based anomaly detection, where adaptive or distribution-aware thresholds are used to balance false alarms and missed detections. Finally, Mahalanobis-based losses or hybrid cosine/MSE objectives can improve robustness when sensor channels are highly correlated, an effect analogous to covariance-aware classifiers such as Linear and Quadratic Discriminant Analysis in supervised learning.

The present study situates itself within this body of work by contrasting a lightweight dense AE against an LSTM-AE, both trained with MSE and, for the dense model, an optional Mahalanobis loss. Thresholds are determined from healthy reconstruction-error distributions, making the pipeline fully data-driven and label-agnostic.

## III. CASE STUDY

**DATA DESCRIPTION**

Figure 1 illustrates the multivariate time series collected from the hydraulic-pump monitoring system. Sensors capture one complete sample every minute, yielding a high-resolution record that spans several months of continuous operation.

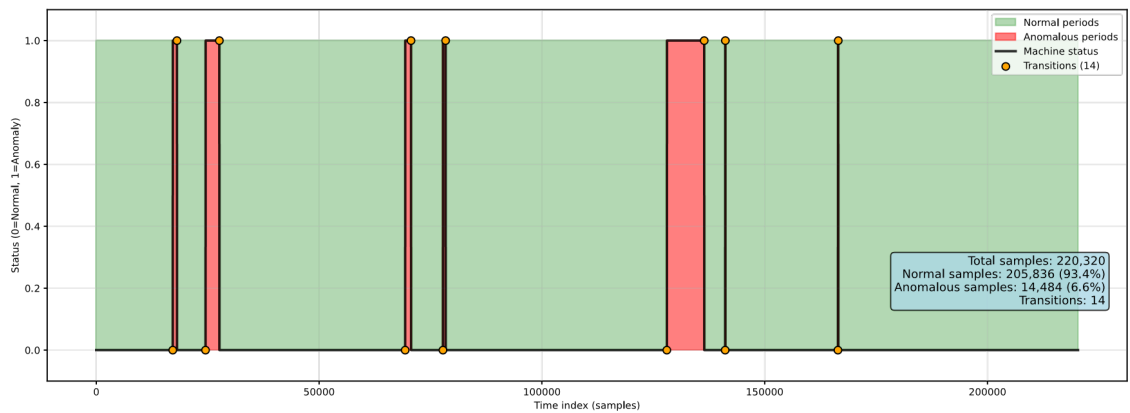

**Figure 1:** Abnormal data distribution.

Each timestamped sample contains 52 variables distributed across four categories: hydraulic parameters (pressures, flow rates, oil temperature), mechanical vibrations (axial, radial and bearing accelerations), electrical measurements (motor current, voltage, power factor), and ambient or supervisory signals (room temperature, load set-points). This diverse set of signals provides a comprehensive view of the pump's mechanical and operational state.

During the observation window, seven fault events were officially annotated by maintenance staff. Table 1 lists their start times and durations. These events vary markedly in length (some last only a few hours, while others persist for several days), posing a significant challenge for anomaly-detection algorithms that must recognise both abrupt and slowly developing deviations.

Table 1: Annotated fault intervals in the hydraulic-pump dataset.

| Fault # | Start (timestamp) | Duration |
|---|---|---|
| 1 | 2018-04-12 21:55 | 15 h 45 min |
| 2 | 2018-04-18 00:30 | 51 h 51 min |
| 3 | 2018-05-19 03:18 | 21 h 53 min |
| 4 | 2018-05-25 00:30 | 10 h 6 min |
| 5 | 2018-06-28 22:00 | 139 h 51 min |
| 6 | 2018-07-08 00:11 | 42 min |
| 7 | 2018-07-25 14:00 | 1 h 16 min |

Healthy operation dominates the dataset; the combined fault intervals account for roughly 1% to 3% of all recorded samples. This severe class imbalance underscores the suitability of unsupervised or one-class approaches that learn a model of normal behaviour instead of relying on numerous fault examples.

**DATA PREPARATION**

To prevent any information leakage, all annotated fault intervals are withheld for evaluation. The remaining fault–free portion of the log is then partitioned into a 90/10 stratified split:

1. *Training set.* The first subset contains 90% of the healthy samples (approximately $1.85 \times 10^5$ observations). These records are used to fit the AEs and, in the Mahalanobis variant, to estimate the residual covariance matrix $\Sigma$.

2. *Test set.* The second subset combines two sources: (i) the remaining 10% of previously unseen normal data and (ii) all anomalous samples. This yields roughly $3.5 \times 10^4$ time steps, of which 42% are labelled as faults.

Table 2: Dataset summary after the 90/10 healthy split.

| Subset | Samples | Anomalies | Features |
|---|---|---|---|
| Train | ∼ 185000 | 0% | 52 |
| Test | ∼ 35000 | 42% | 52 |

This strict separation ensures that model optimisation is driven solely by normal behaviour, while performance metrics are computed on a fully unseen mixture of healthy and faulty observations, thus providing an unbiased assessment of generalisation to real-world conditions.

The raw file exhibits an overall completeness of ∼ 97 %, yet several sensors contain notable gaps. Sensor 15 is completely empty and therefore removed from the dataset. The remaining missing values are handled in a three-step cascade: linear interpolation along the time axis, backward fill to cover leading gaps, and finally, if still required, forward fill to patch trailing gaps. This strategy was specifically chosen because multiple sensors contain null values at different timestamps, making it critical to preserve the temporal coherence of the data, particularly for time-series analysis. The approach preserves local trends while yielding a fully dense matrix suitable for learning, ensuring that the sequential relationships between data points remain intact for downstream temporal modelling.

Once imputation is complete, each of the 51 retained sensor channels is linearly rescaled to the interval [0, 1] using a MinMax transformation fitted exclusively on the healthy training data. The same scaler is then applied to the test set, preventing any information leakage from faulty records into the training pipeline.

For temporal modelling, the normalised stream is segmented into overlapping windows of five consecutive minutes with a stride of one. The dense AE consumes single, window-free snapshots (51 features), whereas the LSTM-AE takes these windows as input sequences. A window is labelled anomalous if at least one of its five timestamps carries a fault flag; otherwise, it is considered normal. This approach ensures that brief excursions are captured while allowing the sequence model to learn the temporal context surrounding each fault.

Finally, 20% of the healthy training samples (individual snapshots for the dense model and windows for the LSTM) are set aside as a validation split. They guide early stopping and learning-rate scheduling, while no fault labels influence model optimisation, guaranteeing an unbiased evaluation.

**EXPLORATORY DATA ANALYSIS**

To obtain an intuitive picture of how normal and faulty samples differ, a three–dimensional t-Distributed Stochastic Neighbor Embedding (t-SNE) projection (Maaten & Hinton, 2008) was generated from a random subset of 5,000 observations. The projection, shown in Figure 2, maps the original 52-dimensional feature space to three coordinates while conserving local neighbourhood structure.

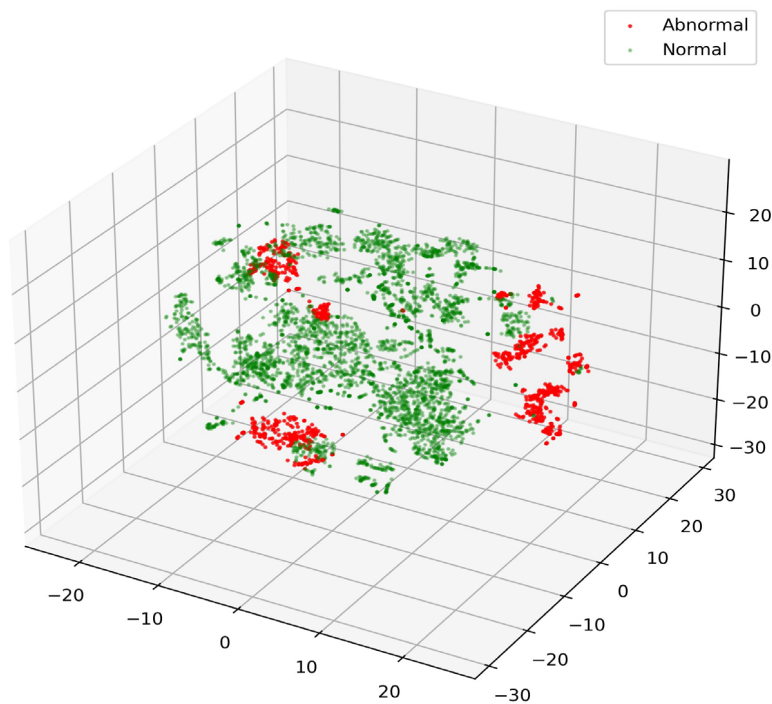

**Figure 2:** Three-dimensional t-SNE embedding of 5 000 points from the dataset.

The embedding reveals a clear separation between operating regimes. Healthy samples (green) form a dense, contiguous manifold that resembles a hollow ring or shell, whereas faulty samples (red) cluster in several compact islands distributed around the periphery. Most anomalous clusters lie well outside the main healthy manifold, suggesting that the underlying sensor patterns during fault conditions are markedly different from normal behaviour. A small degree of overlap can be observed where the red points encroach on the green shell; these points coincide with the early stages of certain faults and indicate that incipient degradation may still share some characteristics with normal data.

The visual separation supports the suitability of reconstruction-error–based detection: if a low-dimensional projection already shows distinct groupings, an AE trained only on healthy samples should produce higher reconstruction errors for the outlying red regions. In addition, the scatter of red clusters highlights that the seven faults are heterogeneous as some clusters are tightly packed, while others are more diffuse, so a flexible detector capable of capturing multiple anomaly modes is required.

Altogether, the t-SNE analysis confirms that meaningful structure exists in the data: normal operation occupies a well-defined region of feature space, whereas anomalies deviate consistently, providing a solid foundation for unsupervised learning.

**FEATURE ENGINEERING**

No manual feature extraction is performed in this study. All 51 normalised sensor channels are fed directly to the AEs, which act as data-driven feature extractors. During training, each network learns a compact latent representation that retains the most relevant information for accurate reconstruction; features that carry little or redundant information are automatically down-weighted or discarded in this process.

*Dense AE.* The feed-forward model maps every 51-dimensional snapshot onto an eight-neuron bottleneck. This latent vector captures nonlinear correlations among hydraulic, vibration, and electrical signals in a form that is maximally useful for rebuilding the original input. Because reconstruction error is later used as the anomaly score, any dimension of the latent space that does not contribute to faithful reconstruction naturally receives a near-zero weight, effectively filtering out noise without explicit feature selection.

Figure 3 visualises the learned latent space on the test set after reducing the eight dimensions to two. It can be seen how the two types of data are clearly differentiated, confirming that the AE has isolated the most discriminative features without any handcrafted engineering.

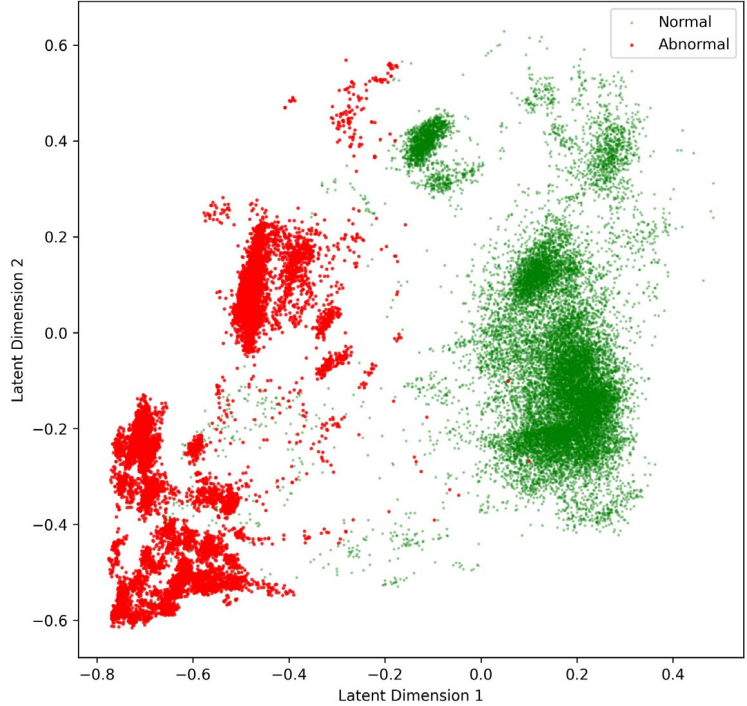

**Figure 3:** Two-dimensional projection of the eight-dimensional latent space for the test set.

*LSTM-AE.* For the sequence model, each input window consists of five consecutive snapshots (5 × 51 values). The encoder compresses this 260-element tensor into the same eight-dimensional latent vector, but now the representation also encapsulates short-term temporal dynamics like subtle trends or oscillations that unfold over the five-minute horizon. By forcing the entire window through a narrow bottleneck, the network learns to preserve those temporal patterns that are characteristic of normal operation and to discard momentary fluctuations that do not matter for reconstruction.

In both cases, the latent space serves as an automatically selected set of features: the encoder decides which combinations of raw signals best summarise the healthy behaviour of the hydraulic pump. This approach eliminates the need for labour-intensive, hand-crafted indicators while ensuring that the most informative aspects of the data drive anomaly detection.

**MODELS**

Two complementary AE structures are employed: a feed-forward (dense) AE and a LSTM-AE. Both are trained in an unsupervised manner on healthy hydraulic-pump data.

*Dense AE.* Let $d = 51$ denote the number of sensor features. The encoder squeezes the $d$-dimensional snapshot through three fully connected layers, while the decoder symmetrically reverses this compression. The complete mapping $f_\theta: \mathbb{R}^d \to \mathbb{R}^d$ is thus:

$$\underbrace{Dense\,(36) \to Dense(12) \to Dense(8)}_{encoder} \to \underbrace{Dense\,(12) \to Dense(36) \to Dense(d)}_{decoder},$$

where every layer uses a *tanh* activation. The eight-neuron bottleneck supplies the latent representation $\mathbf{z} \in \mathbb{R}^8$.

*LSTM-AE.* To capture temporal dynamics, we feed the network with sequences of $T = 5$ consecutive snapshots (shape $T \times d$). The encoder stacks two LSTM layers of sizes 16 and 8, respectively; the decoder mirrors this arrangement after a RepeatVector layer and terminates with a TimeDistributed(Dense($d$)) head. All recurrent units employ *tanh* activations, producing a reconstructed tensor of shape $T \times d$ as follows:

$$\underbrace{LSTM\,(16) \to LSTM(8)}_{encoder} \xrightarrow{RepeatVector} \underbrace{LSTM\,(8) \to LSTM(16) \to TimeDistributed(Dense(d))}_{decoder}.$$

**PROBLEM-SPECIFIC ADAPTATION**

- *Snapshot vs. Sequence Modelling.* The dense AE offers a lightweight baseline that operates on individual time steps, whereas the LSTM-AE leverages sliding windows of $T = 5$ frames to detect gradual temporal drift.
- *Custom Loss Options.* Both networks can be trained with the mean squared error (MSE). The dense AE additionally supports a Mahalanobis loss: first, a short MSE warm-up (five epochs) estimates the residual covariance $\Sigma$ on healthy data; training then proceeds with a loss that accentuates statistically rare error directions. The Mahalanobis loss is defined as:

$$\mathcal{L}_{Mahalanobis} = \frac{1}{N} \sum_{i=1}^{N} \left\| (\hat{x}_i - x_i) \Sigma^{-1/2} \right\|_2.$$

- *Activation Consistency.* Uniform use of *tanh* ensures bounded gradients and symmetry around zero, aligning well with the min–max scaled input features.

**TRAINING PROTOCOL**

- *Data Split.* Only healthy records are used for training; 20% of them form a validation set.
- *Optimisation.* Both models use the Adam optimiser. The dense AE trains for up to 25 epochs with learning rate $\eta = 3 \times 10^{-3}$ and batch size 256, while the LSTM-AE adopts the default Adam learning rate with identical batch settings.
- *Regularisation.* An EarlyStopping callback (patience = 10) restores the best weights, and a ReduceLROnPlateau callback lowers the learning rate by a factor of 0.2 after 5 stagnant epochs.
- *Leakage Prevention.* Covariance estimation, threshold tuning, and all hyperparameter choices rely strictly on training data, preventing any information leak from labelled fault samples.

This design yields a low-overhead dense AE suitable for embedded deployment and a sequence-aware LSTM-AE capable of recognising subtle fault precursors, together offering complementary tools for real-time hydraulic-pump monitoring.

## IV. EVALUATION

**ANOMALY DETECTION MECHANISM**

This section presents two alternative strategies for anomaly detection using AE architectures, evaluating whether incorporating temporal information through sequence modelling improves detection performance compared to non-temporal approaches. The temporal method employs an LSTM-AE processing sliding windows (Malhotra et al., 2016) $x_t \in \mathbb{R}^{s \times w}$ of $s$ sensors over $w$ time steps, while the non-temporal approach uses a standard AE operating on individual observations $x_t \in \mathbb{R}^t$. Both methods compute anomaly scores from reconstruction errors, but differ in their handling of sequential dependencies.

For the temporal approach, the window-wise MSE and corresponding threshold are:

$$MSE_t = \frac{1}{s\,w} \sum_{i=1}^{s} \sum_{j=1}^{w} (x_{t,i,j} - \hat{x}_{t,i,j})^2, \quad \tau_{seq} = \text{Percentile}_\alpha(\{MSE_t^{\text{train}}\}).$$

The pointwise detection method offers two scoring alternatives: (i) standard MSE (Lachekhab et al., 2024) and (ii) Mahalanobis distance (Denouden et al., 2018), with respective thresholds:

$$MSE_t = \frac{1}{d}\sum_{i=1}^{d}(x_{t,i} - \hat{x}_{t,i})^2, \quad \tau_{\text{static}} = \text{Percentile}_\alpha(\{MSE_t^{\text{train}}\}),$$

$$D_t = \sqrt{(\boldsymbol{x}_t - \hat{\boldsymbol{x}}_t)^\top \boldsymbol{\Sigma}^{-1} (\boldsymbol{x}_t - \hat{\boldsymbol{x}}_t)}, \quad \tau_{\text{MD}} = \text{Percentile}_\alpha(\{D_t^{\text{train}}\}),$$

where $\boldsymbol{\Sigma}$ is the residual covariance matrix estimated from training data. The temporal method explicitly captures evolving patterns through sequential processing, while the pointwise approach provides computational simplicity when temporal context is unavailable or unnecessary. Both methods maintain strict separation between training (healthy data only) and evaluation phases to prevent information leakage.

**EVALUATION PROTOCOL**

Unlike many real-world prognostics datasets, the hydraulic pump dataset used in this study provides explicit labels for each sample, indicating whether the system is operating normally or abnormally. This allows for a straightforward and rigorous evaluation of the anomaly detection models.

After computing reconstruction errors on the test set, we compare the model's binary predictions (based on a fixed anomaly threshold) against the ground-truth labels to populate the confusion matrix: True Positives (TP), False Positives (FP), True Negatives (TN), False Negatives (FN). From these counts, we derive standard classification metrics: These metrics jointly capture the model's ability to identify faults early (high recall) while minimising false alarms (high precision and specificity).

In the context of many anomaly detection problems, recall is especially critical, as the primary objective is to detect all potential faults early to avoid unexpected breakdowns (Hundman et al., 2018). However, precision also remains important to reduce false alarms and avoid unnecessary maintenance interventions. Therefore, while the evaluation emphasises sensitivity, achieving a reasonable trade-off with precision is essential for practical deployment.

## V. RESULTS

The results in Table 3 reveal that all three models deliver exceptionally high detection quality. Every metric reported (precision, recall, specificity, and $F_1$) exceeds 0.93, indicating industrial-grade reliability for hydraulic-pump monitoring. Sensitivity (recall) is the foremost concern in this application, because an undetected fault can lead to costly unplanned downtime. Both AEs trained with MSE loss (AE + MSE and LSTM-AE + MSE) achieve a recall of 0.997, while the Mahalanobis-based variant pushes recall to an almost perfect 0.999. In other words, virtually every true anomaly in the test set is detected.

**Table 3:** Detection performance.

| Model | Precision | Recall | Specificity | F1-score |
|---|---|---|---|---|
| AE (MSE) | 0.933 | 0.997 | 0.949 | 0.964 |
| AE (Mahalanobis) | 0.934 | 0.999 | 0.951 | 0.966 |
| LSTM-AE (MSE) | 0.935 | 0.997 | 0.951 | 0.965 |

Maintaining such high sensitivity usually raises the risk of false alarms, yet precision remains around 0.93–0.94 for all approaches. Concretely, roughly 93% of issued alerts correspond to genuine faults, so maintenance crews are seldom dispatched unnecessarily. Specificity, ranging from 0.949 to 0.951, confirms that fewer than 5% of healthy samples are mistakenly flagged, constituting an acceptably low nuisance rate given the large class imbalance in real operation. The harmonic mean of precision and recall ($F_1$ between 0.964 and 0.966) summarises this favourable trade-off, with the Mahalanobis variant enjoying a slight overall edge.

Despite the high overall precision and recall, we examined the few mistakes the models made. False negatives were exceedingly rare (only a handful of fault samples were missed out of thousands); these misses tended to occur at the very fringes of fault intervals. In other words, in a couple of instances, a fault's earliest or latest moments produced only a slight deviation that stayed below the anomaly threshold, causing a brief miss. Such boundary cases did not substantially threaten the system's safety, since the fault was detected shortly after; however, they highlight the inherent trade-off in threshold setting. A slightly lower threshold could have caught those edge points at the cost of more false alarms. The Mahalanobis AE's virtually zero false negatives (recall 0.999) indicate that its distance-based scoring succeeded in flagging even those mild initial deviations, whereas the pure MSE-based models skipped them. This is likely because the Mahalanobis metric inflates anomaly scores for unusual error patterns (even if the absolute error is small) by accounting for correlations in the residual vector. On the false positive side, each model produced a small number of false alarms (false positive rate $\approx 5\%$). We reviewed these cases and found no persistent pattern indicating pathological behaviour; the false alarms appeared sporadically, often on days when the pump was operating under conditions not well represented in the training data (e.g. unusual ambient temperature spikes or brief transients during system start-ups/shut-downs). Such contexts momentarily pushed one or more sensor readings outside the learned normal manifold, triggering an alarm. This phenomenon is a known challenge in unsupervised anomaly detection: the models sometimes flag novel but benign conditions as anomalies simply because they lie outside prior experience.

From an operational standpoint, the AE + Mahalanobis approach is the safest choice when the overriding objective is to avoid any missed fault. The LSTM-AE achieves almost identical aggregate performance while inherently modelling temporal drift, which could offer extra robustness as usage patterns evolve. The plain AE + MSE meets all performance targets with the smallest computational footprint, making it attractive for embedded or low-power edge deployment. Hence, the final selection should reflect practical constraints like computational budget, expected data complexity, and the relative cost of false positives versus missed detections, rather than raw accuracy alone.

## VI. CONCLUSIONS AND FUTURE WORK

*Conclusions.*

High recall is paramount in predictive-maintenance scenarios: missing even a single fault can translate into costly unplanned downtime. In our study, the point-wise dense AE reached the highest recall while, at the same time, demanding the least computational effort for both training and inference. Although the LSTM-AE can, in principle, capture slow temporal drifts, achieving comparable performance required heavier preprocessing (windowing, sequence padding) and longer hyperparameter tuning, with no tangible gain in recall on this dataset.

Therefore, for the concrete use case of minute-level hydraulic-pump monitoring, the snapshot-based AE strikes the best balance between reliability and deployment cost: it attains industrial-grade accuracy, saturates recall close to 1.0, and is lightweight enough for real-time execution on embedded hardware. Temporal models remain valuable when the fault signature is known to unfold over longer horizons, but they should be adopted only when their added complexity is justified by a measurable boost in detection sensitivity.

Finally, an unsupervised learning strategy is particularly well suited to this domain. Hydraulic-pump datasets are dominated by healthy operation, contain few labelled failures, and exhibit heterogeneous fault modes that evolve with machine age. By learning only from normal data, AEs sidestep the need for costly fault annotation, adapt naturally to unseen failure patterns, and improve as more healthy data are logged. This makes the unsupervised dense AE not just the most accurate and efficient option for the present study but also a robust, future-proof foundation for continuous condition monitoring in high-stakes industrial environments.

*Future work.*

Several aspects limit the immediate deployment of the proposed framework. First, the evaluation relies on offline processing; real-time inference may require dedicated edge hardware or model compression. Second, window length and stride in the LSTM-AE were fixed heuristically. A systematic study of sequence granularity, perhaps via Bayesian optimization, could uncover more optimal settings. Third, the ground-truth labels mark distinct fault intervals but do not specify the precise onset of degradation. More granular annotations would enable lead-time analysis and finer calibration of early-warning thresholds. Fourth, the approach treats each sensor equally; incorporating domain knowledge to weight or group sensors could improve interpretability and performance. Finally, explainability techniques such as SHAP, Layer-wise Relevance Propagation, or counterfactual reconstruction could help maintenance engineers understand why an alert is raised, fostering trust in an unsupervised system.

Future research should therefore explore adaptive window lengths, on-device inference, dynamic thresholding, and explainable latent-space visualisation. A hybrid model combining the AE with a physics-informed digital twin of the pump could further enhance reliability under previously unseen operating regimes.

## ACKNOWLEDGEMENTS

This work has been carried out within the framework of the "Cátedra ENIA IA[3]: Cátedra de Inteligencia Artificial en Aeronáutica y Aeroespacio", subsidised by the "Ministerio de Asuntos Económicos y Transformación Digital" (Secretaría de Estado de Digitalización e Inteligencia Artificial), del Gobierno de España.

## VII. REFERENCES

Ahmad, S., Styp-Rekowski, K., Nedelkoski, S., & Kao, O. (2020). Autoencoder-based condition monitoring and anomaly detection method for rotating machines. 2020 IEEE International Conference on Big Data (Big Data), 4093–4102.

Carrasco, J., López, D., Aguilera-Martos, I., García-Gil, D., Markova, I., Garcia-Barzana, M., Arias-Rodil, M., Luengo, J., & Herrera, F. (2021). Anomaly detection in predictive maintenance: A new evaluation framework for temporal unsupervised anomaly detection algorithms. Neurocomputing, 462, 440–452.

Carvalho, T. P., Soares, F. A., Vita, R., Francisco, R. d. P., Basto, J. P., & Alcalá, S. G. (2019). A systematic literature review of machine learning methods applied to predictive maintenance. Computers & Industrial Engineering, 137, 106024.

Denouden, T., Salay, R., Czarnecki, K., Abdelzad, V., Phan, B., & Vernekar, S. (2018). Improving reconstruction autoencoder out-of-distribution detection with mahalanobis distance. arXiv preprint arXiv:1812.02765.

Díaz Álvarez, Y., Hidalgo Reyes, M. Á., Lagunes Barradas, V., Pichardo Lagunas, O., & Martínez Seis, B. (2022). A hybrid methodology based on CRISP-DM and TDSP for the execution of preprocessing tasks in Mexican environmental laws. In O. Pichardo Lagunas, J. Martínez-Miranda, & B. Martínez Seis (Eds.), Advances in Computational Intelligence. MICAI 2022. Lecture Notes in Computer Science (Vol. 13613). Springer, Cham. https://doi.org/10.1007/978-3-031-19496-2_6

Fic, P., Czornik, A., & Rosikowski, P. (2023). Anomaly detection for hydraulic power units—a case study. Future Internet, 15(6). https://doi.org/10.3390/fi15060206

Guo, Y., Liao, W., Wang, Q., Yu, L., Ji, T., & Li, P. (2018). Multidimensional time series anomaly detection: A gru-based gaussian mixture variational autoencoder approach. Asian conference on machine learning, 97–112.

Haakman, M. (2020). AI life cycle models need to be revised: An exploratory study in fintech. https://doi.org/10.48550/arxiv.2010.02716

Herrmann, L., Bieber, M., Verhagen, W. J., Cosson, F., & Santos, B. F. (2024). Unmasking overestimation: A reevaluation of deep anomaly detection in spacecraft telemetry. CEAS Space Journal, 16(2), 225–237.

Hundman, K., Constantinou, V., Laporte, C., Colwell, I., & Soderstrom, T. (2018). Detecting spacecraft anomalies using lstms and nonparametric dynamic thresholding. Proceedings of the 24th ACM SIGKDD international conference on knowledge discovery & data mining, 387–395.

Ionescu, A., Mouw, Z., Aivaloglou, E., Katsifodimos, A., Fekete, J., Omidvar-Tehrani, B., Rong, K., & Shraga, R. (2024). Key insights from a feature discovery user study. Proceedings of the 2024 Workshop on Human-In-the-Loop Data Analytics, 1–5.

Jakubowski, J., Stanisz, P., Bobek, S., & Nalepa, G. J. (2021). Anomaly detection in asset degradation process using variational autoencoder and explanations. Sensors, 22(1), 291.

Lachekhab, F., Benzaoui, M., Tadjer, S. A., Bensmaine, A., & Hamma, H. (2024). Lstm-autoencoder deep learning model for anomaly detection in electric motor. Energies, 17(10), 2340.

Maaten, L. v. d., & Hinton, G. (2008). Visualizing data using t-sne. Journal of machine learning research, 9(Nov), 2579–2605.

Malhotra, P., Ramakrishnan, A., Anand, G., Vig, L., Agarwal, P., & Shroff, G. (2016). Lstm-based encoder-decoder for multi-sensor anomaly detection. arXiv preprint arXiv:1607.00148.

Martínez-Plumed, F., et al. (2021). CRISP-DM twenty years later: From data mining processes to data science trajectories. IEEE Transactions on Knowledge and Data Engineering, 33(8), 3048–3061. https://doi.org/10.1109/TKDE.2019.2962680

Moratilla, A., Fernández, E., Álvarez, A., & Narciso, A. (2023a). Prediction of the number of days for sale in the real estate asset market. International Journal of Modern Research in Engineering and Technology (IJMRET), 8(10).

Moratilla, A., Fernández, E., Álvarez, A., & Narciso, A. (2023b). Strategic locations of electric vehicle charging facilities utilizing mixed-integer linear programming and genetic algorithm models. International Journal of Modern Research in Engineering and Technology (IJMRET), 8(10).

Oakes, B., Famelis, M., & Sahraoui, H. (2024). Building domain-specific machine learning workflows: A conceptual framework for the state of the practice. ACM Transactions on Software Engineering and Methodology, 33, 41–50. https://doi.org/10.1145/3638243

Qian, J., & Song, S. (2022). A review on autoencoder-based representation learning for fault detection. Chemometrics and Intelligent Laboratory Systems, 224, 104548. https://doi.org/10.1016/j.chemolab.2022.104548

Rollins, J. (2015). Foundational methodology for data science. Domino Data Lab, Inc., Whitepaper.

Said Elsayed, M., Le-Khac, N.-A., Dev, S., & Jurcut, A. D. (2020). Network anomaly detection using lstm based autoencoder. Proceedings of the 16th ACM symposium on QoS and security for wireless and mobile networks, 37–45.

Sakurada, M., & Yairi, T. (2014). Anomaly detection using autoencoders with nonlinear dimensionality reduction. MLSDA.

Schmetz, A., Kampker, A., et al. (2024). Inside production data science: Exploring the main tasks of data scientists in production environments. AI, 5(2), 873–886. https://doi.org/10.3390/ai5020043

Shearer, C. (2000). The CRISP-DM model: The new blueprint for data mining. Journal of Data Warehousing, 5(4), 13–22.

Taslim, D., & Murwantara, I. (2024). Comparative analysis of ARIMA and LSTM for predicting fluctuating time series data. Bulletin of Electrical Engineering and Informatics.

Tziolas, T., Papageorgiou, K., Theodosiou, T., Papageorgiou, E., Mastos, T., & Papadopoulos, A. (2022). Autoencoders for anomaly detection in an industrial multivariate time series dataset. Engineering Proceedings, 18(1). https://doi.org/10.3390/engproc2022018023

Zenati, H., Romain, M., Foo, C.-S., Lecouat, B., & Chandrasekhar, V. (2018). Adversarially learned anomaly detection. 2018 IEEE International conference on data mining (ICDM), 727–736.